\documentclass{article}

\usepackage{arxiv}
\usepackage{subcaption}
\usepackage[utf8]{inputenc} 
\usepackage[T1]{fontenc}    
\usepackage{hyperref}       
\usepackage{url}            
\usepackage{booktabs}       
\usepackage{amsfonts}       
\usepackage{nicefrac}       
\usepackage{microtype}      
\usepackage{lipsum}
\usepackage{amsmath,amsfonts}
\usepackage{xcolor}
\usepackage{pifont}  

\usepackage{graphicx}
\graphicspath{ {./images/} }
\usepackage[utf8]{inputenc} 
\usepackage[T1]{fontenc}    
\usepackage{hyperref}       
\usepackage{url}            
\usepackage{booktabs}       
\usepackage{amsfonts}       
\usepackage{nicefrac}       
\usepackage{microtype}      
\usepackage{lipsum}
\usepackage{graphicx}
\graphicspath{ {./images/} }
\usepackage{xcolor}
\usepackage{amsmath}
\usepackage{tcolorbox}
\definecolor{turqoise}{RGB}{79, 225, 201}

\title{Human-in-the-loop reasoning for traffic sign detection: Collaborative approach YOLO with Video-LLaVA}

\author{
Mehdi Azarafza \\
Department of Computer Science \\
 Univ. of Applied Science Hamm-Lippstadt \\
  Lippstadt, Germany\\
  \texttt{mehdi.azarafza@hshl.de} \\
   \And
Fatima Idrees\\
Department of Computer Science \\
 Univ. of Applied Science Hamm-Lippstadt \\
  Lippstadt, Germany\\
  \texttt{fatima.idrees@hshl.de} \\
   \And
Ali Ehteshami Bejnordi\\
Department of Computer Science \\
 Univ. of Applied Science Hamm-Lippstadt \\
  Lippstadt, Germany\\
  \texttt{ali.ehteshami-bejnordi@hshl.de} \\
  \And
 Charles Steinmetz \\
Department of Computer Science \\
 Univ. of Applied Science Hamm-Lippstadt \\
  Carl von Ossietzky University of Oldenburg\\
  Lippstadt, Germany\\
  \texttt{charles.steinmetz@hshl.de} \\
  \And
 Stefan Henkler \\
Department of Computer Science \\
 Univ. of Applied Science Hamm-Lippstadt \\
  Lippstadt, Germany\\
  \texttt{stefan.henkler@hshl.de} \\
  \And
 Achim Rettberg \\
Department of Computer Science \\
 Univ. of Applied Science Hamm-Lippstadt\\
 Carl von Ossietzky University of Oldenburg\\
  Lippstadt, Germany\\
  \texttt{achim.rettberg@hshl.de}  \\
}

\begin{document}
\maketitle
\begin{abstract}
Traffic Sign Recognition (TSR) detection is a crucial component of autonomous vehicles. While You Only Look Once (YOLO) is a popular real-time object detection algorithm, factors like training data quality and adverse weather conditions (e.g., heavy rain) can lead to detection failures. These failures can be particularly dangerous when visual similarities between objects exist, such as mistaking a 30 km/h sign for a higher speed limit sign. This paper proposes a method that combines video analysis and reasoning, prompting with a human-in-the-loop guide large vision model to improve YOLO's accuracy in detecting road speed limit signs, especially in semi-real-world conditions. It is hypothesized that the guided prompting and reasoning abilities of Video-LLava can enhance YOLO's traffic sign detection capabilities. This hypothesis is supported by an evaluation based on human-annotated accuracy metrics within a dataset of recorded videos from the CARLA car simulator. The results demonstrate that a collaborative approach combining YOLO with Video-LLava and reasoning can effectively address challenging situations such as heavy rain and overcast conditions that hinder YOLO's detection capabilities.
\end{abstract}


\section{Introduction}
The deployment of Intelligent Transportation Systems (ITS) is experiencing a significant rise in the paradigm of smart cities. One of the key components of ITS is traffic object detection, this technology leverages computer vision and image processing to identify specific objects (e.g., vehicles, pedestrians) within digital imagery.\cite{jiao2019}.
Traffic Sign Recognition (TSR) is one of the most crucial aspects in Autonomous Vehicles (AVs)\cite{nur2021}, especially for interpreting speed limit signs that reflect the maximum permissible speed on the road. Expanding the diversity of a dataset used to train the YOLO algorithm, such as incorporating synthetic images alongside real-world images, has been shown to positively impact its object recognition performance.\cite{dewi2021yolo}. However, the performance of YOLO can be influenced by weather conditions. 

\begin{figure}

\centerline{\includegraphics[height=6.5cm]{ 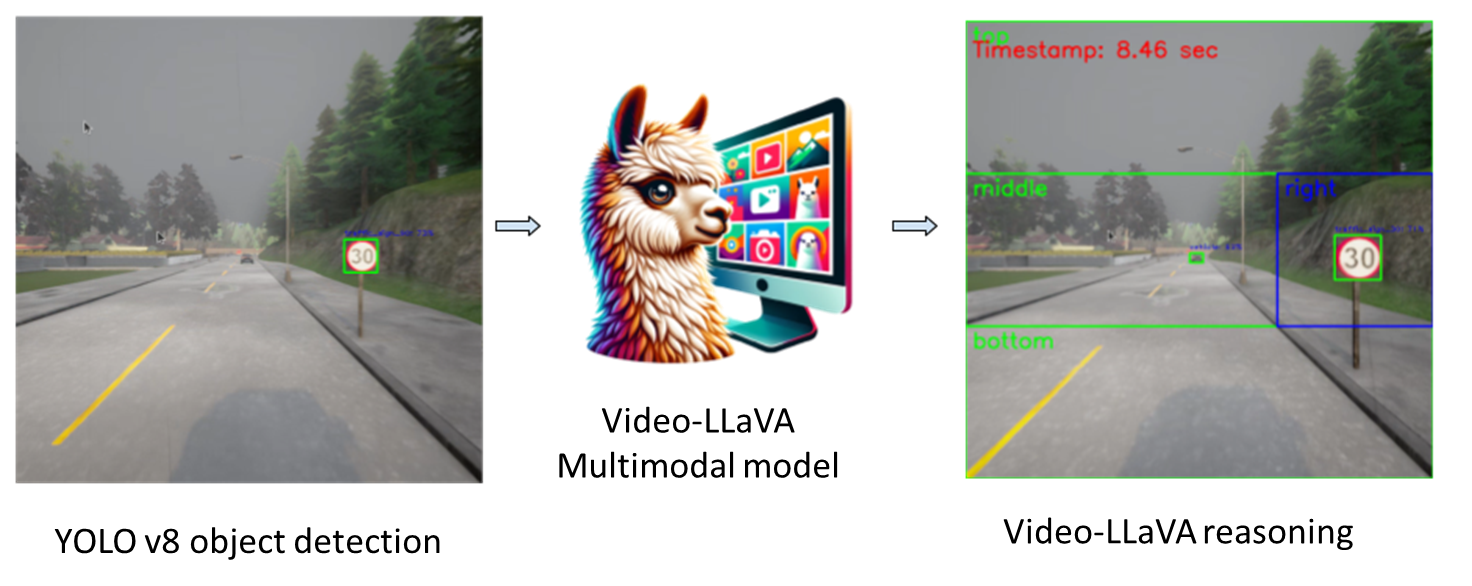}}
\caption{High level overview of reasoning with llava-video}
\label{fig:arch}
\end{figure}

To address this challenge under semi-real-world conditions, this paper proposes a novel method that leverages iterative reasoning with Video-LLaVA\cite{lin2023video} to enhance the accuracy and reliability of YOLO's object detection, specifically for speed limit traffic signs. The process begins with a recorded video analyzed by YOLO v8 to detect speed limit signs and determine their values. A human expert then evaluates YOLO's performance. If YOLO fails to accurately identify the speed limit, its output is fed into Video-LLaVA. The expert prompts Video-LLaVA with general questions, using each response to refine subsequent prompts. This iterative approach helps Video-LLaVA to pinpoint YOLO's inaccuracies and correctly determine the speed limit. 
Figure \ref{fig:arch} illustrates high-level overview reasoning. The recorded video output from YOLO v8 serves as input for Video-LLaVA at this stage. A human expert poses general questions, with each response subsequently becoming input for the next prompt. The human expert does not provide direct answers. Following this process and localizing the inaccuracies from YOLO, Video-LLaVA attempts to determine the correct speed limit.

The rest of the paper is organized as follows. Section 2 provides related work. The large Vision Language Model is described in Section 3. Section 4 provides an overview implementation of object detection with human experts in the loop prompting with Video-LLaVA. The analysis of the experimental scenarios is presented in Section 5. Results and comparison between YOLO v8, Video-LLava direct answer without human and human-in-the-loop(HITL) guidance reasoning are presented in Section 6. Finally, the last section is the study is conclusion. \\

\section{Related work} 

Large Language Models (LLMs) have demonstrated exceptional performance in the domain of Natural Language Processing (NLP) with applications such as news summarization, sentiment analysis, and text-based reasoning. These models have the capability to understand data in the form of text and images, generate textual data, and provide solutions to tasks that require complex reasoning. In  \cite{azarafza2024hybrid} Azarafza et al. (2024) explored hybrid reasoning based on LLMs for autonomous driving scenarios, demonstrating the potential of these models to improve decision-making processes in dynamic environments by analyzing the detected objects and sensor data when provided with additional context.

Video Multi-Modality Models process input videos and generate responses according to the prompts/commands provided by users. At the core of these Multi-Modality Models, the fundamentals of object detection are required for the identification of the region of interest (ROI) within the video while responding to the user prompts. This also covers a larger spectrum of research works on text generation, image captioning, image-video encoding, and more. Recent advancements in object detection have seen the emergence of LLaVA models, which can deal with both visual and text data. Unlike traditional methods, LLaVA models integrate descriptive language information to enhance object detection, classification, and localization.

Recently, it has been demonstrated in Chain-of-Spot by Liu et al. (2024) \cite{liu2024chainofspot} that interactive reasoning can significantly enhance the performance of Large Vision-language Models. This method dynamically identifies and focuses on the key regions of interest (ROIs) within the images, leading to a much improved visual feature extraction and state-of-the-art results on several benchmarks. 

Additionally, Jiao et al. (2024) \cite{jiao2024enhancing} conducted an empirical study demonstrating that integrating state-of-the-art object detection and OCR models with Multimodal Large Language Models (MLLMs) can significantly improve image understanding. They incorporated models like LLaVA-1.5 \cite{liu2023improved}, DINO\cite{caron2021emerging}  and PaddleOCRv2\cite{du2021pp}. The enhanced LLaVA-1.5 achieved up to a 12.5\% improvement on normalized average scores across ten benchmarks, demonstrating the effectiveness of text-based detection information in improving overall performance.

Dang et al. (2024) proposed InstructDET \cite{dang2024instructdet}, a data-centric method for referring object detection (ROD) that allows the localization of target objects based upon the input/instructions provided by the user. By utilizing foundation models to produce human-like object detection instructions, they constructed the InDET dataset, which encompasses a wide range of user instructions related to object detection. This method and use of InDET dataset demonstrated significant improvements over existing methods on standard REC datasets and the newly developed InDET set.

Another relevant work by Pan et al. (2024) \cite{pan2024monitoring} involved exploring the use of YOLOv8 combined with LLaVA for monitoring human activities in riverine spaces. This study demonstrated that alongside the fundamental counting and locating of individuals, the fusion of these models with the implementation of guided prompts enabled the recognition of human activities like walking, running, and skateboarding. Thus, enhancing the environmental understanding significantly. 

The recent works in the Video Multi-Modality Models and Object Detection have demonstrated significant improvements in visual feature extraction and understanding. However, there remains a need to explore the integration of visual and text data to improve performance across a wider range of applications. Therefore, this research aims to analyze the integration of text-based reasoning to guide the large-vision model which will assist YOLO in improving the results of TSR.

\section{Large Vision-Language Model} 
Large language models (LLMs) employ transformer-based neural network architecture and are trained on vast amounts of text data to identify patterns and relationships in natural language. These models have demonstrated the ability to both interpret and produce text like human language. However, LLMs are primarily focused on processing text inputs and outputs, lacking the ability to directly perceive and reason about visual information like images and videos. This limitation has motivated research into extending LLMs to handle multi-modal inputs beyond just text. Vision-language models (VLMs), also known as large vision-language models (LVLMs), aim to bridge this gap by enabling LLMs to understand and generate responses based on both textual and visual inputs. These models typically involve integrating computer vision components like convolutional neural networks (CNNs) or vision transformers (ViTs) \cite{dosovitskiy2020image} to extract visual features, which are then aligned and fused with the text representations within the LLM architecture. Approaches like pre-training on large image-text or video-text datasets and instruction tuning \cite{liu2024visual} have shown promising results in grounding LLMs in visual contexts for tasks like visual question answering, image/video captioning, and multimodal reasoning. Recent visual-language models like Video-LLaVA \cite{lin2023video} take this a step further by introducing a unified approach to visual-language modeling, integrating both image and video data into a single, coherent representation for large language models. Video-LLaVA is based on an auto-regressive language model, which is fine-tuned on multimodal instruction-following data generated by Llava1.5 \cite{liu2023improved} and VideoChat\cite{li2023videochat}.

\begin{figure}

\centerline{\includegraphics[height=10.5cm]{ 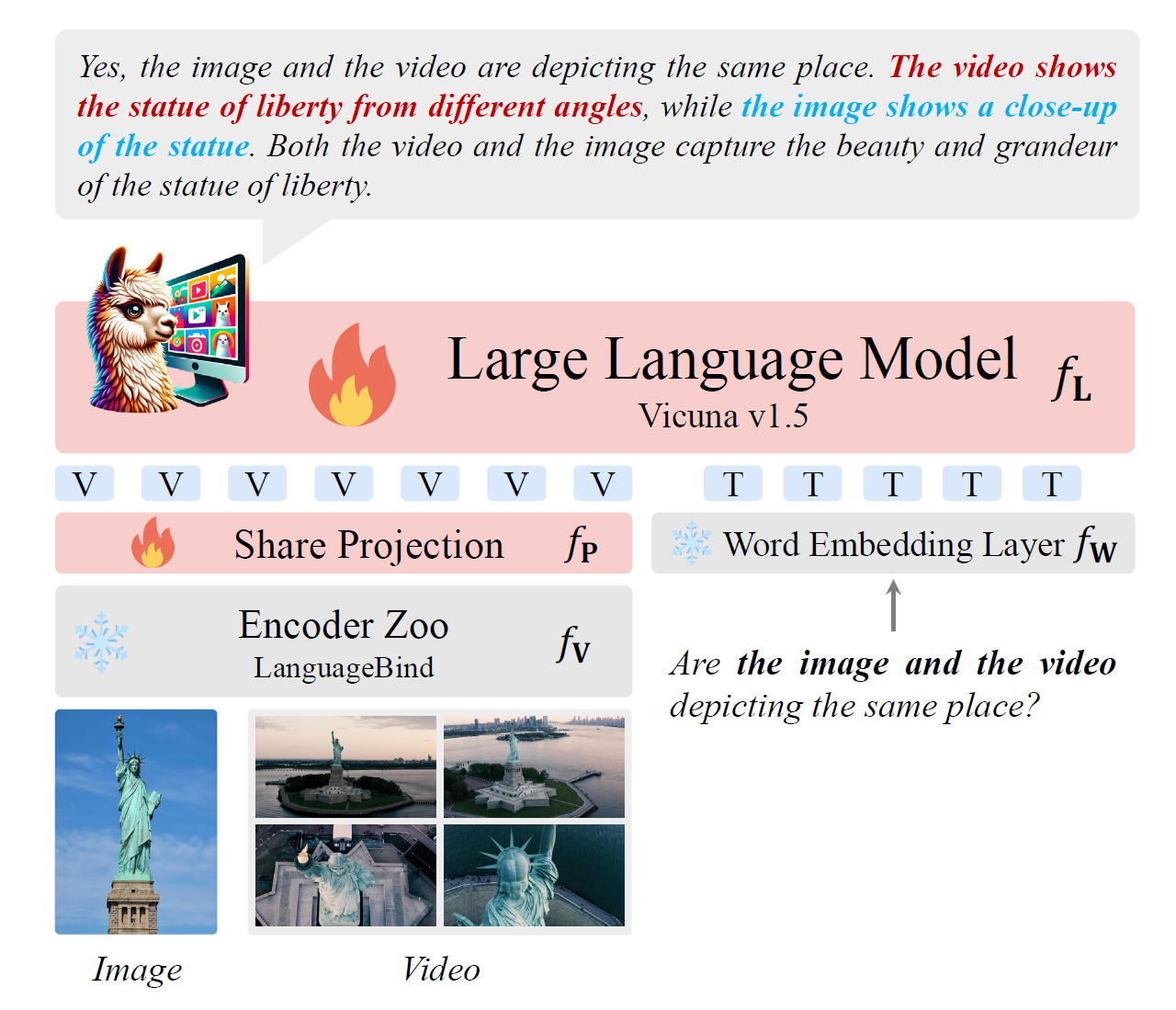}}
\caption{Video-LLaVA structure \cite{lin2023video}  }
\label{fig:llava}
\end{figure}

As shown in Figure \ref{fig:llava} the structure of Video-LLaVA is designed to effectively integrate visual and textual data into a unified representation for enhanced multi-modal reasoning. Central to its structure are the LanguageBind encoders \cite{zhu2023languagebind}, which extract features from both images and videos and map these visual signals into a shared textual feature space, facilitating a unified visual representation. These encoders are initialized from OpenCLIP \cite{cherti2023reproducible}, ensuring that the visual features are aligned with the language features, thereby creating a unified visual representation. The core of the model is the Large Language Model (LLM), specifically Vicuna-7B v1.5 \cite{zheng2024judging}, which processes this unified visual representation along with tokenized textual queries to generate coherent and contextually relevant responses. In order to facilitate this integration, visual projection layers are employed to encode the unified visual representation before it is combined with the textual queries. Finally, the Word Embedding Layer tokenizes the textual input, converting it into a format that can be processed by the LLM.

The training involves encoding textual and visual signals into token sequences and maximizing the likelihood probability to achieve multi-modal understanding. The training pipeline involves two main stages: understanding training and instruction tuning. In the understanding training stage, the model learns to interpret visual signals using a large dataset of image/video-text pairs, focusing on auto-regressive loss to develop basic visual comprehension. Instruction tuning then enhances the model's ability to respond to complex visual comprehension tasks based on different instructions, incorporating multiple rounds of conversation data.

\section{Object detection reasoning with Video-LLaVA} 
\begin{figure}[htbp]
\centerline{\includegraphics[ height=12cm]{ 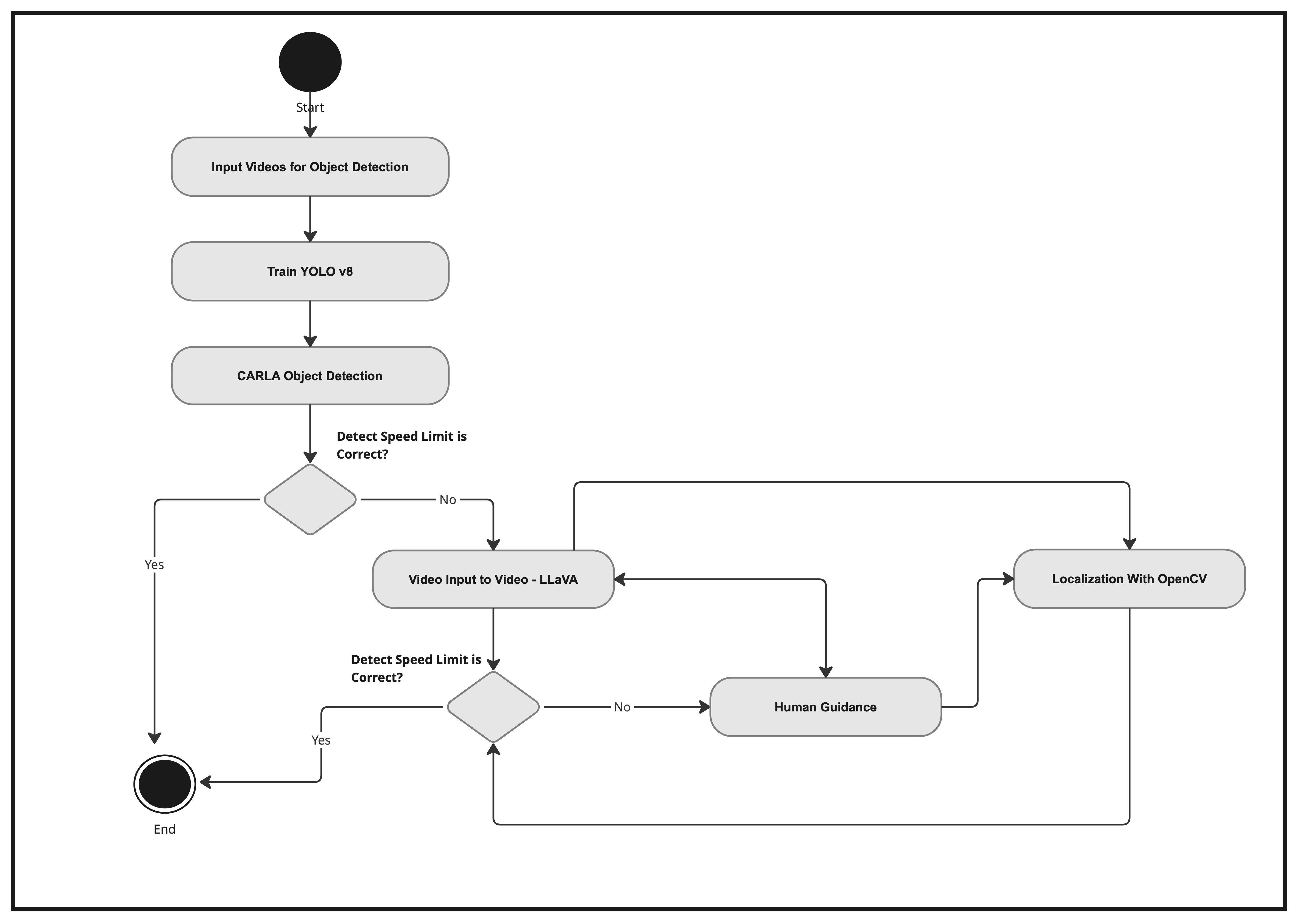}}
\caption{Activity Diagram for the Collaborative Approach of Human-in-the-Loop Reasoning Using YOLO and Video-LLaVA}
\label{fig:flow}
\end{figure}

This section presents the details of the the workflow associated with the proposed human-in-the-loop reasoning and prompting a collaboration between YOLO and Video-LLaVA\cite{lin2023video} for speed limit recognition. Figure \ref{fig:flow} presents the activity diagram of this approach. The process begins with inputting data and training YOLO v8, one of the state-of-the-art models from the YOLO family, within the CARLA car simulator. The focus here is on detecting speed limits on the road; other objects like vehicles, traffic lights, bicycles, and pedestrians are excluded. Video recordings are captured iteratively.\\ If YOLO correctly identifies the speed limit, the process is complete. Otherwise, the video serves as input for Video-LLaVA. If Video-LLaVA can detect the speed limit on the first attempt, the workflow is finished. Otherwise, a human expert intervenes to provide guidance based on Video-LLaVA's initial response. It is crucial to emphasize that human prompts are formulated iteratively based on Video-LLaVA's own outputs. In each step, Video-LLaVA's response serves as the foundation for the subsequent human prompt.\\ \\
The human expert's role is to provide guidance that refines the focus area by considering spatial limitations. This iterative process continues until the correct speed limit is identified. It is important to underscore that human intervention does not involve directly providing the answer. This collaborative approach enhances the system’s reasoning capabilities, focusing on critical areas until the correct speed limit is detected.\\

\section{Analysis of scenarios}  
This section presents an analysis of two showcase scenarios and output of YOLO, Video-LLaVA, and a reasoning approach that leverages Video-LLaVA to improve upon YOLO's outputs. Twenty videos were randomly recorded, and the evaluation is based on these 20 short videos, each ranging from 3 to 6 seconds in length.\\ The focus of the object detection evaluation and analysis is on 'speed limit' signs, while other detected objects are not considered. 
\
As a showcase of the 20 recorded videos, 2 videos were randomly selected. These scenarios are implemented within the CARLA simulator to showcase the performance of these methods. The scenarios are implemented within the CARLA simulator, with the ego vehicle set to autopilot mode. Town02 is chosen as the city environment, where 25 vehicles are randomly spawned. To increase the difficulty of detection, weather conditions are set to 'heavy rain' and 'fully cloudy'. 
\begin{figure}[ht]
  \subcaptionbox*{a) YOLO v8: Detected Speed Limit}[.50\linewidth]{%
    \includegraphics[width=\linewidth,height=9cm]{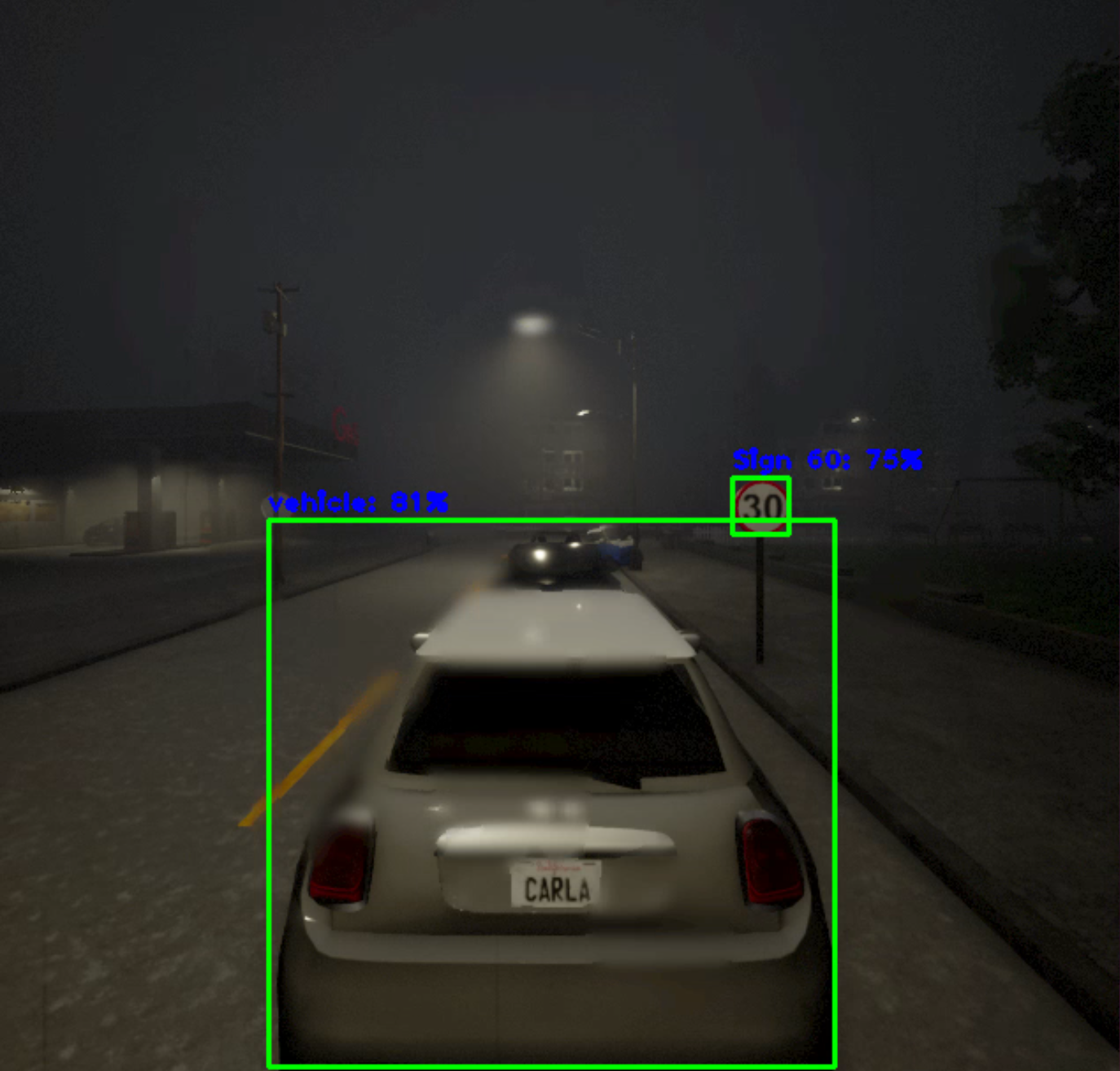}%
  }%
  \hfill
  \subcaptionbox*{b) human-in-the-loop reasoning with Video-LLaVA: Detected Speed Limit Location}[.50\linewidth]{%
    \includegraphics[width=\linewidth,height=9cm]{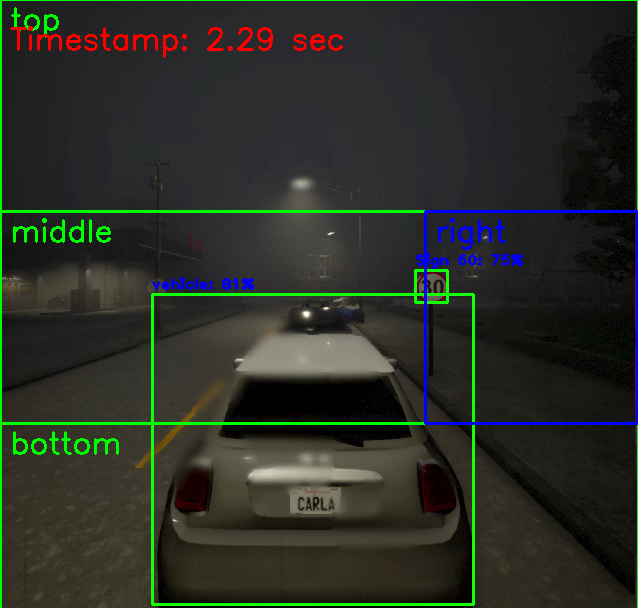}%
  }
  \caption{a) YOLO Output: Speed Limit Sign Roads 30, Detected as 60 (Blue Font 'Sign 60'). b) Step-by-Step human-in-the-loop with Video-LLaVA: Refining Detection with Regioning}
  \label{fig:scenario1}
\end{figure}

\vspace{-10px}

\subsection{Scenario 1}
The first scenario employs a single randomly selected video clip. The ego vehicle is in motion, and a single "speed limit 30" traffic sign is positioned in the road's right corner. Figure \ref{fig:scenario1} showcases the outputs for YOLO v8 and the human-in-the-loop reasoning approach. As depicted in Figure \ref{fig:scenario1}a), YOLO successfully detects both the vehicle and the speed limit sign. However, the detected speed limit via YOLO is incorrectly identified as "60." In contrast, Figure \ref{fig:scenario1}b) demonstrates that Video-LLaVA, guided by human input through four questions without providing the exact answer, was able to correctly determine the speed limit as 30. The prompts used to guide the human-in-the-loop reasoning process are presented within a dedicated highlighted box. It is important to note, that the primary focus here is on rectifying incorrect detections and leveraging Vision Language Models (VLMs) to enhance YOLO's performance. It's important to acknowledge that YOLO exhibits numerous accurate detections that are not considered within the scope of this scenario.

\begin{tcolorbox}[
  colback=white!1,  
  colframe=black!100, 
  arc=5mm,           
  parbox=false,     
  title=Human-in-the-Loop reasoning for Video-LLava scenario 1,
]


 Prompt 1: what is this video about? \\\\
Answer: This video is about a car driving down a street at night, with a green square on the side of the car. \\
\rule{\linewidth}{0.4pt} 


 Prompt 2: This video is about a simulated car driving down a street at night, with a green square on the side of the car indicating a detected object. What is the speed limit on the road? \\\\
 Answer: The speed limit on the road is 35.  \textcolor{red}{\fontsize{12}{12}\selectfont \ding{55}} (incorrect)

 \rule{\linewidth}{0.4pt} 

Prompt 3:  This video is about a simulated car driving down a street at night, with a green square on the side of the car indicating a detected object. Is there a speed limit sign in this city? If so, where is it located in the driver's view: on the right, middle, or left?\\\\

Answer: Yes, there is a speed limit sign in this city. It is located on the right side of the driver's view.\\\\
 \rule{\linewidth}{0.4pt} 
Prompt 3: This video is about a simulated car driving down a street at night, with a green square on the side of the car indicating a detected object. What is the speed limit on the road? \\\\
Answer:  The speed limit in this city is 30.
 \textcolor{green}{\fontsize{12}{12}\selectfont \textbf{\ding{51}}} (correct).

\end{tcolorbox}

\subsection{Scenario 2}
\begin{figure}[ht]
  \subcaptionbox*{a) YOLO v8: Undetected Speed Limit}[.50\linewidth]{%
    \includegraphics[width=\linewidth,height=8.5cm]{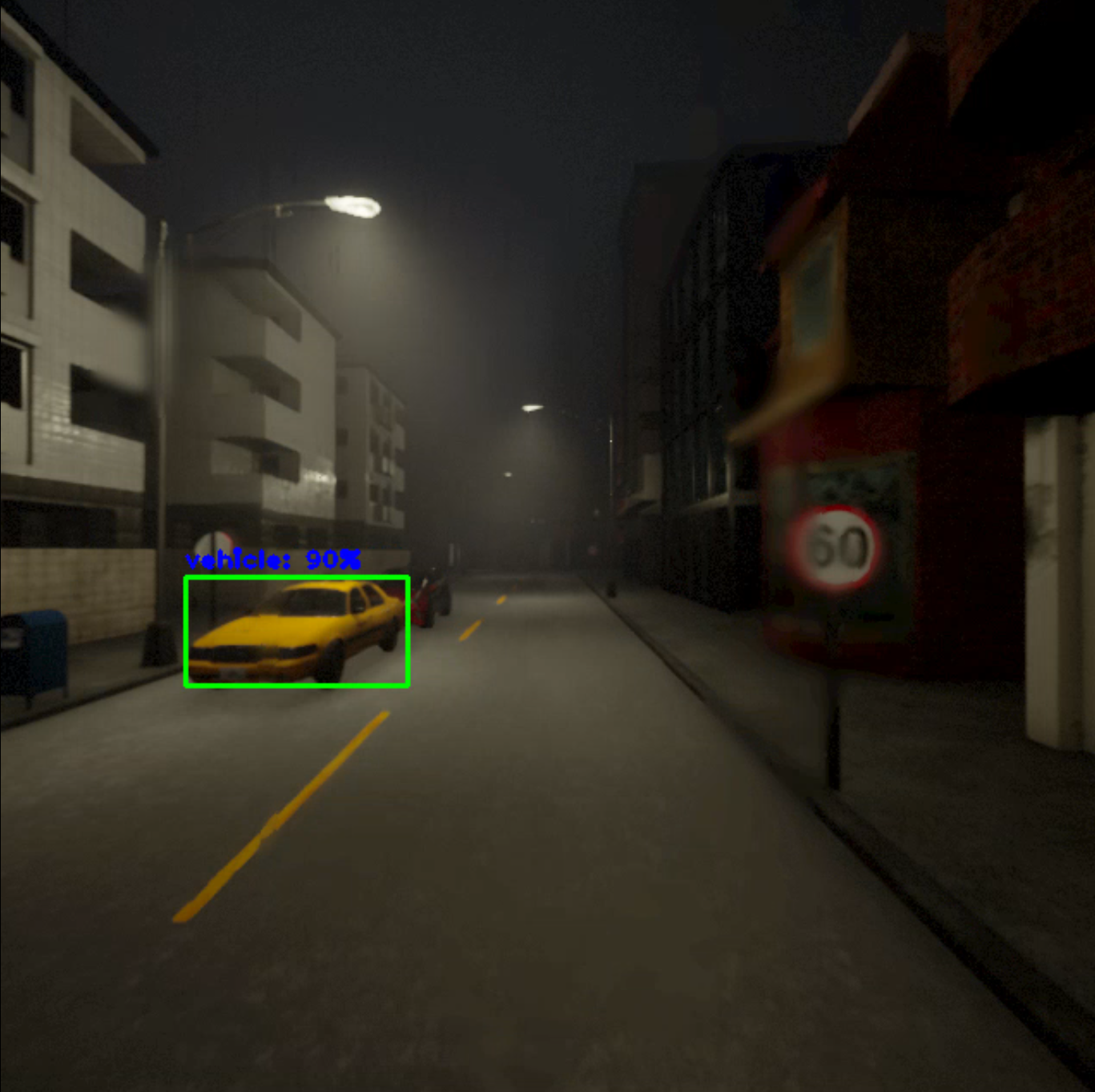}%
  }%
  \hfill
  \subcaptionbox*{b) human-in-the-loop reasoning with Video-LLaVA: Detected Speed Limit Location}[.50\linewidth]{%
    \includegraphics[width=\linewidth,height=8.5cm]{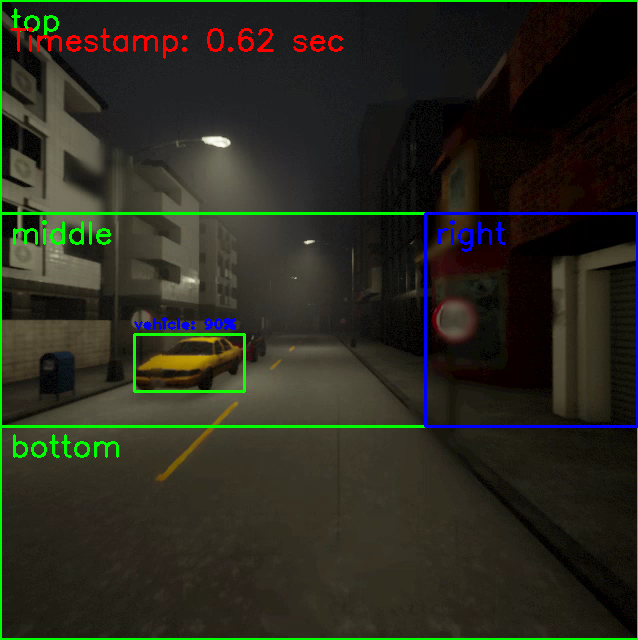}%
  }
  \caption{a) YOLO Output: Initial Failure to Detect Speed Limit Due to Heavy Rain Weather Condition. b) Step-by-Step human-in-the-loop with Video-LLaVA: Refining Detection with Regioning for guide detection}
  \label{fig:scenario2}
\end{figure}

The second scenario features the ego vehicle crossing the city environment in "heavy rainy" weather conditions, similar to Scenario 1. However, a key distinction lies in the potential impact of the rain on camera functionality, as illustrated in Figure \ref{fig:scenario2}. Consequently, YOLO initially fails to detect the object in this scenario. This scenario exemplifies a challenging situation for camera systems, as adverse weather conditions can significantly hinder object detection capabilities. \\This scenario highlights a limitation of YOLO in performing object detection under blurred conditions. While vehicle detection is successful, the speed limit sign remains undetected for the first few seconds when it is blurred. On the other hand, Video-LLava's initial prompt, a general question asking "What is this video about," receives an approximately accurate response. However, the response mentions the virtual world (CARLA) but incorrectly states that "the player is in control of a yellow car." The yellow car is the vehicle that approached from the front, not the ego vehicle.   Despite this error, it's important to acknowledge that Video-LLava demonstrates the ability to detect the speed limit sign, unlike YOLO which experiences initial detection failures. \\
The third prompt leverages human expertise to refine the initial prompt generated by Video-LLaVA. In this phase, corrections are directed toward mitigating potential errors identified within the previous Video-LLaVA output. Furthermore, human guidance incorporates the concept of a region of interest (ROI) to focus attention on the specific location of the traffic sign within the video frame. \\
The fourth prompt builds upon the information provided in the previous prompt's response. In this iteration, the answer incorporates more detailed explanations, as exemplified by the accurate statement "the street is lined with buildings" derived from the video content. Additionally, the answer identifies the speed of the ego vehicle, a detail not previously mentioned by human input. Notably, this prompt successfully guides Video-LLaVA to correctly detect the speed limit as 60 km/h within the city environment. It is important to acknowledge that Video-LLaVA persists in a minor error, classifying the ego vehicle as yellow. However, this is deemed an acceptable discrepancy as it falls outside the primary focus of the prompt.

\begin{tcolorbox}[
  colback=white!1,  
  colframe=black!100, 
  arc=5mm,           
  parbox=false,     
  title=Human-in-the-Loop reasoning for Video-LLava scenario 2,
]


 Prompt 1: What is this video about? \\
Answer: This video is about a game of "driving" in a virtual world where the player is in control of a yellow car. The game is played in a city at night, and the player is required to follow traffic rules and avoid obstacles. 
\rule{\linewidth}{0.4pt} 


 Prompt 2: This video is about a game where the player drives a car in a virtual world. The game is set in a city at night. What is the speed limit on the road?\\
 Answer: The speed limit on the road is 30.  \textcolor{red}{\fontsize{12}{12}\selectfont \ding{55}} (incorrect)

\rule{\linewidth}{0.4pt} 
Prompt 3: This video is about a CARLA car simulator where the ego vehicle drives a car in a virtual world. The environment is set in a city at night and in heavy rainy conditions. Is there a speed limit sign in this city? If so, where is it located in the driver's view: on the right, middle, or left? \\\\
Answer: Yes, there is a speed limit sign in the driver's view. It is located on the right side of the road. \\\\
\rule{\linewidth}{0.4pt} 
Prompt 4: This video is about a CARLA car simulator where the ego vehicle drives a car in a virtual world. The environment is set in a city at night and in heavy rainy conditions. There is a speed limit sign in this city, located on the right side of the driver's view. What is the speed limit?\\\\
Answer:  The video shows a yellow car driving down a street at night. The car is moving at a speed of 30 km/h, and there is a speed limit sign on the right side of the driver's view. The street is lined with buildings, and the car is driving on the road with a speed limit of \colorbox{turqoise}{60} km/h.
 \textcolor{green}{\fontsize{12}{12}\selectfont \textbf{\ding{51}}} (correct). \\The video also shows a green car driving down the same street, but it is not clear if it is the same car as the yellow car.

\end{tcolorbox}

\section{Evaluation}

\begin{figure}[ht]
  \subcaptionbox*{a) Percentage of correct and wrong answer}[.50\linewidth]{%
    \includegraphics[width=\linewidth]{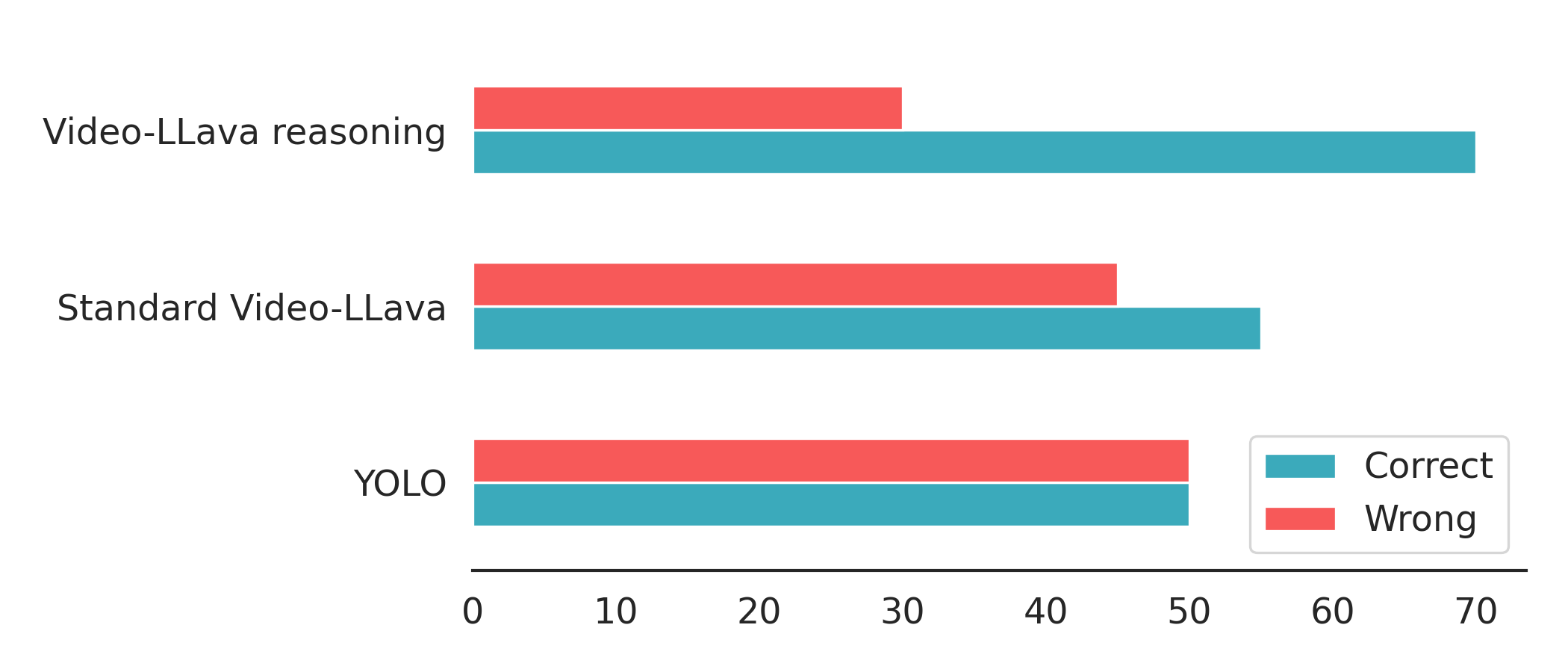}%
  }%
  \hfill
  \subcaptionbox*{b) Accuracy for Compared Methods}[.50\linewidth]{%
    \includegraphics[width=\linewidth]{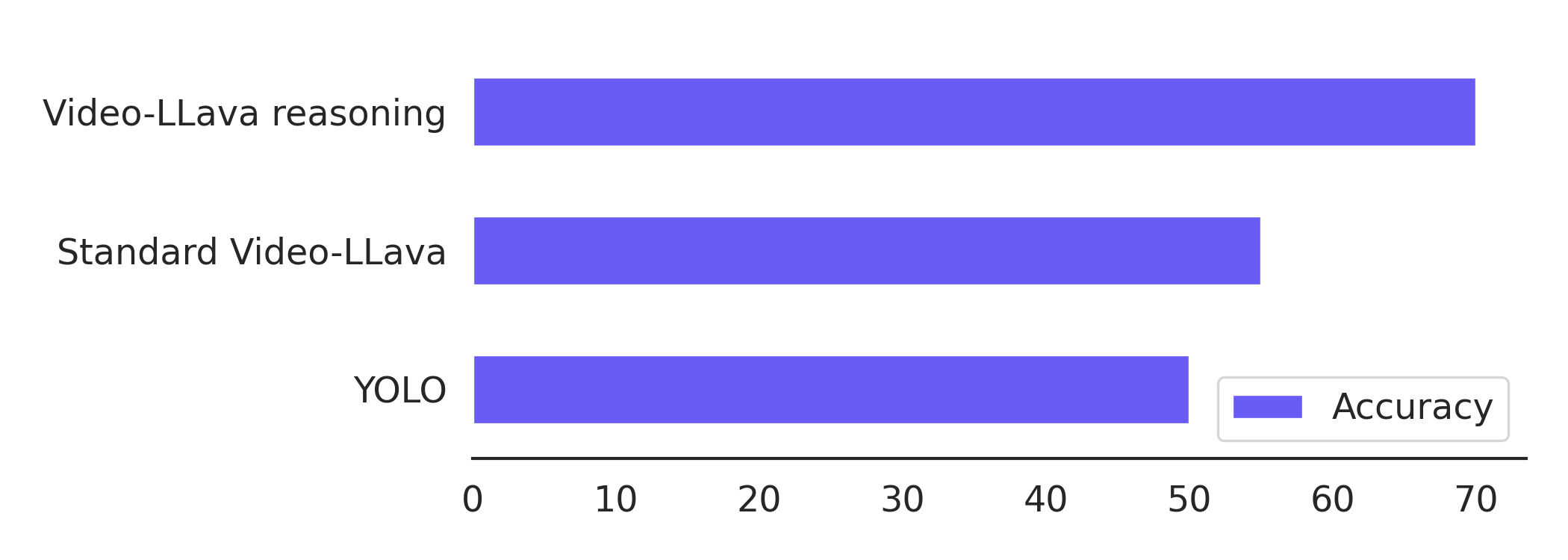}%
  }
  \caption{Comparison of Accuracy and Percentage of Correct and Incorrect Answers Using Three Methods}
  \label{fig:compare}
\end{figure}

To evaluate the performance of the proposed reasoning with a human-in-the-loop approach for correcting mistakes made by YOLO, 20 video outputs from YOLO v8 were randomly recorded. To establish a ground truth for evaluation, a human expert carefully calculated the number of correctly detected speed limit signs within each video. The primary metric employed in this section to assess the performance of the compared methods is accuracy. Accuracy is a fundamental metric for classification tasks, and it is calculated as:
\begin{equation}
\text{Accuracy} = \frac{\text{Number of Correct Predictions}}{\text{Total Number of Predictions}}
\end{equation}
The figure \ref{fig:compare} illustrates a comparative evaluation of three object detection approaches used in this study: Video-LLava reasoning, standard Video-LLava, and YOLO. The 70\% accuracy rate signifies a significant improvement over Standard Video-LLava (55\%) and suggests that incorporating reasoning and prompting techniques can enhance the model's ability to address challenging scenarios or visual ambiguities that might hinder other models. While YOLO offers a baseline performance (50\%), Video-LLava prompting demonstrates a substantial improvement in accuracy, suggesting its potential for real-world applications. 

\section{Conclusion}
This study investigated the integration of textual reasoning to guide a large vision model (YOLO) for improved Traffic Sign Recognition (TSR), particularly focusing on speed limit detection. The research evaluated the performance of YOLO v8 in conjunction with Video-LLava, a system incorporating human-guided prompts. Human-annotated ground truth data served as the basis for accuracy measurement. The results demonstrably show that Video-LLava augmented with reasoning and prompting techniques achieves significantly higher accuracy compared to both standard Video-LLava and the YOLO object detection algorithm. The results showed that reasoning to guide a large-vision model showed a noteworthy improvement over the standard Video-LLaVA, indicating that interactive features can help refine predictions under certain circumstances, leading to better overall performance. Both Video-LLaVA models show a significant number of correct predictions when the speed limit is 30, which could imply specific strengths or biases towards this condition. Further investigation might be necessary to improve their performance in other scenarios.
%
\section{Acknowledgment}
This paper is the result of preliminary work by Hamm-Lippstadt University of Applied Sciences, Germany on the
project ‘Requirements and Application of GAIA-X in the Edge-Device Automobile’ (GAIA-X 4 AGEDA), cosponsored by the German Federal Ministry for Economic Affairs and Climate Action (BMWK 19S22004O) and the
project ‘AutoDevSafeOps’ (ADSO) co-sponsored by the German Federal Ministry of Education and Research (BMBF
011S22087G) and the project 'AKI4KMU - Automatisiertes KI-Framework für KMU' supported by the Ministry of Economic Affairs, Industry, Climate Action and Energy of the State of North Rhine-Westphalia. Co-funded by the European Union. Grant number EFRE-20800498. The sole responsibility for the content lies with the authors.\\
This paper has been accepted at 8th Future of Information and Communication Conference (2025).

\end{document}